\documentclass[runningheads]{llncs}
\usepackage[T1]{fontenc}
\usepackage{graphicx}
\usepackage{hyperref}
\usepackage{color}

\usepackage{amsmath,amsfonts,latexsym,amssymb,euscript,xr}
\usepackage{booktabs}
\usepackage[nodayofweek]{datetime}
\usepackage{float}
\usepackage[table]{xcolor}
\usepackage{multirow}
\usepackage{color,colortbl,tabularx}
\usepackage{amsmath,amssymb}
\usepackage{enumitem}

\usepackage{makecell}
\usepackage{verbatim}
\usepackage[english]{babel}
\usepackage[protrusion=true,expansion=true]{microtype}
\usepackage{amsmath,amsfonts}
\usepackage{pifont}
\usepackage{cleveref}
\usepackage{subcaption,graphicx}

\definecolor{LightBlue}{rgb}{0.88,0.9,0.9}

\def\methodName{M\textsuperscript{3}-Gen}

\begin{document}
\title{\methodName{}: Interpretable Multimodal Generation of Gene Expression Profiles Using Clinical and Imaging Data}
\titlerunning{\methodName{}: Multimodal Generation of Gene Expression Profiles}
% If the paper title is too long for the running head, you can set
% an abbreviated paper title here
%
\author{Francesca Pia Panaccione\inst{1}\orcidID{0009-0005-8007-963X} \and
Carlo Sgaravatti\inst{1}\orcidID{0009-0001-4962-5365} \and
Marco Venere\inst{1}\orcidID{0009-0002-8991-1443}}
\authorrunning{F.P. Panaccione et al.}
% First names are abbreviated in the running head.
% If there are more than two authors, 'et al.' is used.
%
\institute{DEIB - Dipartimento Elettronica, Informazione e Bioingegneria, Politecnico di Milano, Milan, 20133, Italy, 
\email{\{francescapia.panaccione,carlo.sgaravatti,marco.venere\}@polimi.it}}
\maketitle              % typeset the header of the contribution
\begin{abstract}
Integrating heterogeneous biomedical data—including clinical metadata, histopathology images, and molecular profiles—is crucial for comprehensive disease understanding. However, gene expression data acquisition remains constrained by high costs and privacy concerns, limiting its use in multimodal research and AI-driven applications. 
We present MultiModal Molecular Generation (\methodName{}), a novel framework for the generation of gene expression profiles by conditioning a Generative Adversarial Network on histopathology images and clinical metadata. \methodName{} learns a unified latent representation from the clinical variables and the images, leveraging contrastive learning, and exploits the embeddings of the two modalities to guide a generative model in producing biologically coherent gene expression profiles. Evaluations on the TCGA dataset demonstrate that \methodName{} generates realistic and functionally meaningful gene expression data.
Importantly, by integrating multiple modalities in an attention-based mechanism, \methodName{} provides intrinsic explainability: it allows the identification of which regions of the histopathology images most strongly influenced the generation of specific gene expression profiles, making the model’s decisions interpretable by design. Code will be available at: \url{https://github.com/CarloSgaravatti/M3-Gen}.
\end{abstract}

\section{\bf Introduction}
\label{sec:SCIENTIFIC-BACKGROUND}

Biomedical research increasingly relies on integrating heterogeneous data types—\ molecular profiles, clinical metadata, and medical images—to build richer models of biological systems and disease mechanisms. In practice, however, these modalities are unevenly available: while clinical records and histopathology slides are routinely collected and standardized, gene expression profiles remain costly, privacy-sensitive, and limited in scale \cite{liu2024contrastive}. This imbalance not only constrains the scope of multimodal studies but also prevents researchers from fully exploiting the rich contextual information contained in clinical and morphological data. 

Generative AI—from Generative Adversarial Networks (GANs) to denoising diffusion models—presents a promising solution. By learning complex, high-dimensional data distributions, these methods can produce realistic synthetic transcriptomic profiles that augment scarce cohorts and safeguard patient privacy \cite{van2024synthetic}. Despite early successes in generating gene‐expression data (e.g., Panaccione \emph{et al.}~\cite{panaccione2025biogan}, Vinas \emph{et al.} ~\cite{vinas2022adversarial}, Lacan \emph{et al.}~\cite{lacan2023gan}), most approaches rely on limited inputs and overlook critical clinical and histological context.

To overcome these limitations, we introduce \textbf{\methodName{}} (MultiModal Molecular Generation), a unified framework that learns a shared latent representation of clinical variables and histopathology images to produce biologically coherent gene expression profiles \emph{in silico}. To the best of our knowledge, we are the first to address the problem of conditioning a generative model of gene expressions on both images and text. In contrast, existing methods such as \cite{schmauch2020deep,zheng2024digital} focus on directly predicting gene expression from pathology images, without leveraging textual clinical information or generative modeling.
This distinction is important: while prediction is inherently a deterministic task, generative modeling provides a mechanism to explore previously unseen combinations of clinical and histopathological inputs, enabling the generation of multiple plausible omics profiles corresponding to the same conditions.
Furthermore, as demonstrated in this work, \methodName{} inherently supports explainability. By integrating multimodal inputs, the model captures the most relevant visual patterns in the histopathology images in relation to the clinical context when generating a specific gene expression profile. This allows tracing which aspects of the input data contributed most to the output, and observing the correlations and interconnections across the three data modalities (clinical variables, imaging, and gene expression), providing insights into the underlying biological mechanisms. In our parallel line of work, we explored a more complex fusion architecture aimed at increasing representational capacity \cite{gemm-gan}. In contrast, the present manuscript deliberately focuses on interpretability through an attention-based fusion mechanism, enabling patch-level analysis of histopathology images in the gene expression generation process.
We evaluate \methodName{} using standard metrics from the generative modeling literature—such as distributional alignment and downstream predictive performance—to demonstrate its ability to produce realistic, functionally meaningful gene expression profiles. 

\section{\bf Data and Methods}
\label{sec:DATA-AND-METHODS}

\begin{figure}[h]
    \centering
    \begin{subfigure}[t]{0.49\textwidth}
            \begin{center}
                \includegraphics[width=\textwidth]{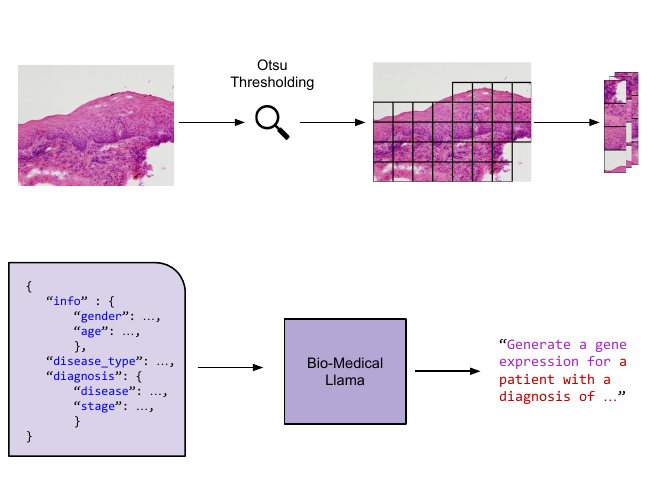}
            \caption{Preprocessing}
            \label{fig:method:preprocessing}
            \end{center}
        \end{subfigure}
        \hfill
        \begin{subfigure}[t]{0.49\textwidth}
            \begin{center}
            \includegraphics[width=\textwidth]{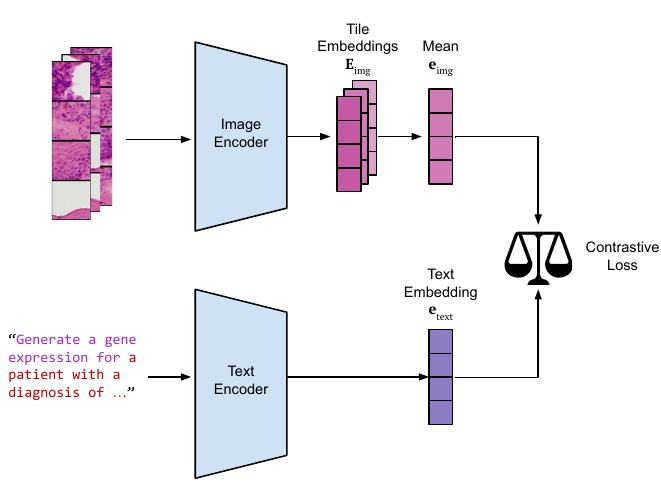}
            \caption{Contrastive Pretraining}
            \label{fig:method:contrastive}
            \end{center}
        \end{subfigure}
      \vspace{0.5em}
        \begin{subfigure}[b]{\textwidth}
        \centering
        \includegraphics[width=\textwidth]{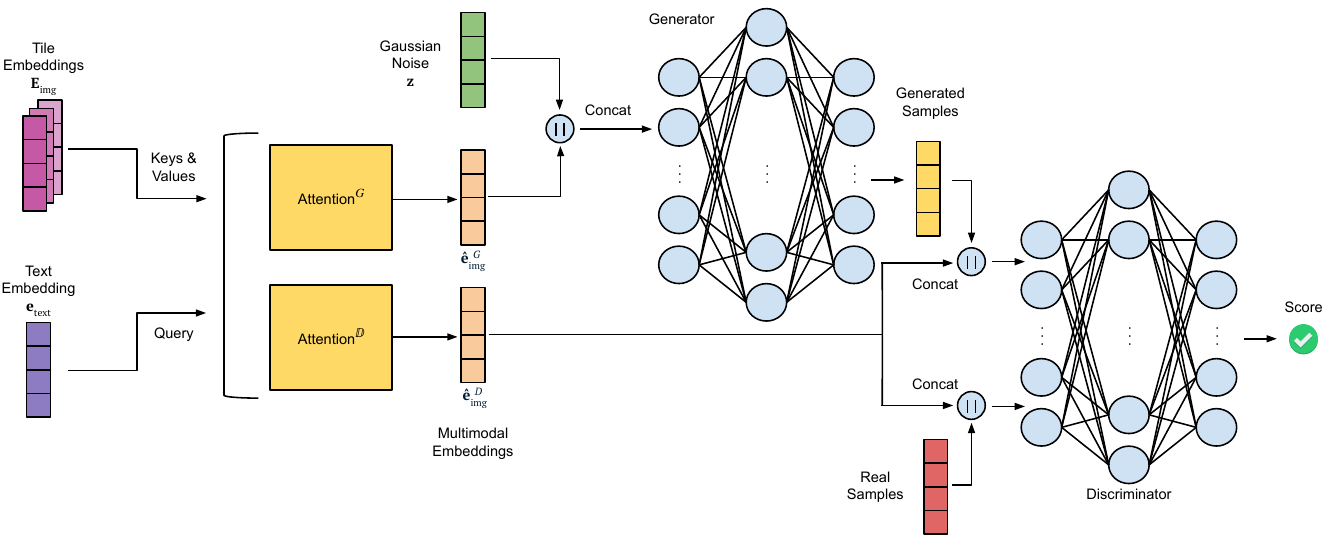}
        \caption{Conditional generation with WGAN-GP}
        \label{fig:method:generative-model}
    \end{subfigure}
    \caption{The overall pipeline of \methodName{}. In the preprocessing step (a), we extract patches from tissue slides using the Otsu thresholding algorithm to select the patches with a percentage of background pixels higher than 80\%, and we extract textual descriptions from patients' metadata with an LLM. We then pre-train an image encoder and a text encoder to align their embeddings with contrastive learning (b). Finally, an attention mechanism is used to combine the embeddings of the patches and of the text to condition a WGAN-GP (c).}
    \label{fig:method}
\end{figure}

At a high level, M\textsuperscript{3}-Gen consists of three main stages: (i) a \emph{preprocessing} phase, where we extract and filter image patches from tissue slides and summarize clinical metadata into compact textual descriptions; (ii) a \emph{contrastive pretraining} step, where we align visual and textual representations in a shared embedding space; and (iii) a \emph{generative model}, where a Wasserstein GANs with Gradient Penalty (WGAN-GP)~\cite{gulrajani2017} is conditioned on these multimodal embeddings to synthesize gene expression profiles. As illustrated in \Cref{fig:method:contrastive}, an \emph{Image Encoder} embeds patches of tissue slides, whose mean representation is aligned, via contrastive learning, with the embedding produced by a \emph{Text Encoder} from the clinical description. These embeddings are then integrated through an attention mechanism to produce a single multimodal embedding. Finally, this multimodal embedding is used to condition the WGAN-GP, by concatenating it to the input noise of the generator and to the real or generated gene expression in input to the discriminator.

\subsection{Data}

The input data for our method consist of paired histopathology whole-slide images (WSIs), clinical metadata, and matched gene expression profiles, all retrieved from TCGA public repository ~\footnote{\url{https://www.cancer.gov/ccg/research/genome-sequencing/tcga}} and focused on twelve different tumor types. WSIs are ultra-high-resolution images of tissue sections, which are computationally infeasible to analyze in full, and are therefore typically subdivided into smaller tiles for downstream processing. Clinical metadata includes patient-specific information such as demographics, cancer subtype, and treatment history, all related to the disease condition. Gene expression profiles, obtained from RNA sequencing, are quantified using FPKM (Fragments Per Kilobase of transcript per Million mapped reads), providing normalized measures of gene activity specific to the disease context.

\subsection{Preprocessing}

To prepare data for multimodal generation, we segment tissue regions from pathology slide thumbnails using Otsu Thresholding \cite{otsu1975threshold} and extract multi-scale high-resolution tiles (128–1024 px), keeping only those tiles with more than 20\% of tissue content. For clinical context, we convert structured metadata into concise case summaries using a quantized instruction-tuned language model. We used a version of Llama3-8B fine-tuned on medical data~\footnote{\url{https://huggingface.co/ContactDoctor/Bio-Medical-Llama-3-8B}} \cite{ContactDoctor_Bio-Medical-Llama-3-8B}. Irrelevant fields are removed, and the remaining data are serialized into prompts, resulting in ~200-word descriptions that capture disease site, demographics, and experimental conditions.

\subsection{Constrastive Pretraining}

To enable effective multimodal conditioning without the need for end-to-end training of the entire model, we first align visual and textual representations in a shared embedding space using contrastive pretraining. 
Contrastive learning is a technique that aims to bring the embeddings of semantically similar data points from different modalities (e.g., an image and its corresponding text) closer together in a shared $d$-dimensional latent space, while pushing unrelated pairs further apart.
Since image patches and clinical descriptions originate from different modalities but are correlated to the same patient, we learn modality-specific encoders with semantically aligned representations.

We fed as input to the Image Encoder a random subset of $N$ patches extracted from the tissue slides, producing as output an embedding matrix $\mathbf{E_{\text{img}}} \in \mathbb{R}^{N \times d}$. The Text Encoder, instead, embeds the clinical textual description into $\mathbf{e_{\text{text}}} \in \mathbb{R}^d$. We then compute the mean $\mathbf{e_{\text{img}}} \in \mathbb{R}^d$ of the patch embeddings and align $\mathbf{e_{\text{text}}}$ and $\mathbf{e_{\text{img}}}$ using the Information Noise-Contrastive Estimation (InfoNCE) loss function, as defined in CLIP \cite{radford2021learning}:
\begin{equation}
\mathcal{L} = \frac{1}{2} \left[ \text{CE}\left( \frac{\text{sim}(\mathbf{e_{\text{img}}}, \mathbf{e_{\text{text}}})}{\tau} \right) + \text{CE}\left(\frac{\text{sim}(\mathbf{e_{\text{text}}}, \mathbf{e_{\text{img}}})}{\tau} \right) \right],
\end{equation}
where $\text{sim}(a, b) = a^\top b$ denotes cosine similarity after $\ell_2$ normalization, and $\tau$ is a temperature hyperparameter. The cross-entropy loss encourages each image to match its paired text, and vice versa.

\subsection{Generative Model}

To synthesize gene expression profiles conditioned on both visual and textual patient information, we employ a \textbf{Conditional WGAN-GP} architecture, currently a standard for transcriptomic data synthesis due to its training stability and capacity to model complex biological distributions. The generator takes as input a latent noise vector  $\mathbf{z} \sim \mathcal{N}(0, I)$ concatenated with a multimodal embedding derived from the contrastive pretraining step. Specifically, we reuse patch-level embeddings $\mathbf{E_{\text{img}}} \in \mathbb{R}^{N \times d}$ and clinical text embeddings $\mathbf{e_{\text{text}}} \in \mathbb{R}^d$. WSIs might contain patches that refer to the diagnosed disease and other patches that are not relevant. Thus, to capture the most relevant visual patterns in relation to the clinical context, we apply a multi-head attention mechanism, using $\mathbf{e_{\text{text}}}$ as the query and $\mathbf{E_{\text{img}}}$ as keys and values.
This produces an attention-weighted image embedding $\mathbf{\hat{e}_{\text{img}}}^G \in \mathbb{R}^d$, which is concatenated with the random latent vector to form the input to the generator. For the discriminator, we apply the same attention mechanism, conditioning on $\mathbf{\hat{e}_{\text{img}}}^D$. The discriminator takes as input both real and generated gene expression profiles and is trained to distinguish whether a sample is real or generated. 

This makes our method inherently interpretable, as we can compute attention maps that can be directly inspected to quantify the contribution of each image patch in conditioning the generated transcriptomic profiles. This provides a biologically meaningful explanation of which tissue regions drive the synthesis process.
Attention maps are directly connected to the attention weights given as output by the multi-head attention layer of the generator. Given a multi-head attention layer with $H$ heads, the weights $\alpha^{(h)} \in \mathbb{R}^N$ for head $h$ are computed as:
\begin{equation}
\boldsymbol{\alpha}^{(h)} = \operatorname{softmax}\!\left(\frac{(W_h^Q \mathbf{e_{\text{text}}})^\top (W_h^K \mathbf{E_{\text{img}}})}{\sqrt{d_h}}\right),
\end{equation}
where $W_h^Q, W_h^K \in \mathbb{R}^{d_h \times d}$ are learnable projection matrices that map, respectively, the textual query $\mathbf{e_{\text{text}}}$ and the visual keys $\mathbf{E_{\text{img}}}$ into a common subspace of dimension $d_h$. To obtain a single interpretable relevance score per patch, we compute the mean of the attention weights across all heads:
\begin{equation}
\boldsymbol{\alpha} = \frac{1}{H} \sum_{h=1}^{H} \boldsymbol{\alpha}^{(h)},
\end{equation}
resulting in $\boldsymbol{\alpha} \in \mathbb{R}^N$, where each element $\alpha_i$ represents the relative importance of the $i$-th patch embedding in conditioning the generator. These weights can be visualized as an attention map over the WSI patches, highlighting the most influential regions for the synthesized gene expression profile.

\section{\bf Experiments}
We test our method on the TCGA dataset, comparing our results with standard generative models. The dataset provides both tissue slides and clinical descriptions of each patient, associated with a gene expression profile derived from RNA-seq. For storage and computational purposes, we select only a subset of TCGA by keeping only the samples having a tissue slide of dimension less than 70 MB, for a total of 1224 clinical cases. 

\subsection{Experimental Setup}
We performed our experiments on a machine with the AMD Ryzen 1950X CPU, with 128 GB of RAM, using two NVIDIA A6000 GPUs with 48 GB of VRAM. We employed PyTorch \footnote{{\url{https://pytorch.org/}}} version 2.6.0 to define our model and perform the experimental evaluation, and CUDA version 12.4 for model training on the GPU devices. 
We use UNI \cite{chen2024uni} as an Image Encoder, which offers a pretrained Vision Transformer for histopathology tiles. As a Text Encoder, we exploit Clinical ModernBERT \cite{lee2025clinical} \footnote{\url{https://huggingface.co/Simonlee711/Clinical_ModernBERT}} \footnote{Both UNI and Clinical ModernBERT are accessible through Huggingface}. We add a linear projection layer to the embeddings of these two models to have the same embedding size for both modalities. In our experiments, we use an embedding size of 128.

\subsection{Evaluation Metrics}
We evaluate our approach with three classes of metrics: (i) \emph{unsupervised metrics} (Precision, Recall \cite{kynkaanniemi2019improved} and Correlation \cite{10.1093/bioinformatics/btad239}), (ii) \emph{detectability}, which consist of training a classifier to detect whether samples are real or generated (lower Accuracy and F1-score means high similarity between real and generated data), and (iii) \emph{utility}, which assesses the performance (Accuracy and F1-score) of models trained on synthetic data and tested on real data to classify the disease type of a patient from RNA-seq.

\subsection{Contrastive Pretraining Results}
\begin{figure}[H]
    \centering
    \includegraphics[width=\linewidth]{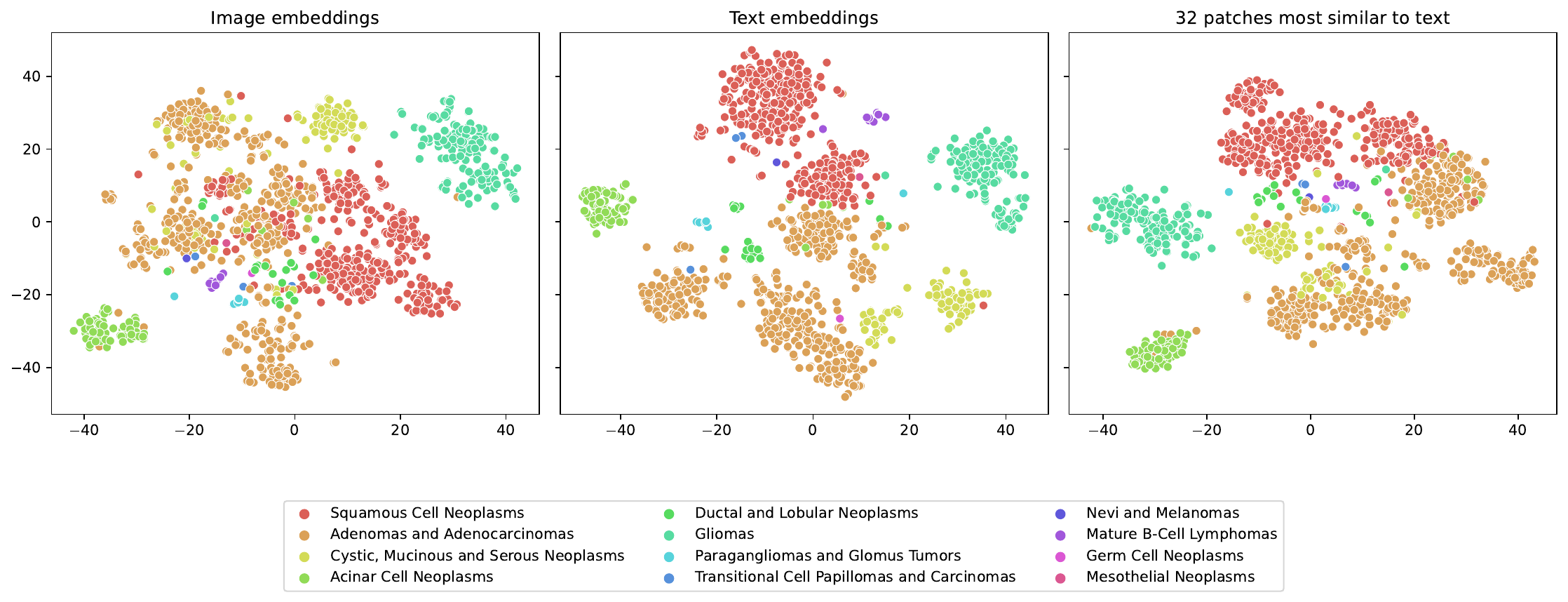}
    \caption{TSNE visualizations of the embeddings after pre-training with contrastive learning. Different colors indicate different disease types.}
    \label{fig:tsne}
\end{figure}

Our feature extractors, UNI and Clinical ModernBERT, are pretrained, respectively, on WSI data and medical descriptions. Thus, to align their representations with contrastive learning, we freeze the pretrained layers and train only the linear projection. At each training step we select 16 random patches to be fed as input to the Image Encoder. We train the model for 20 epochs using a batch size of 8 and the Adam optimizer, with a learning rate of $0.0001$.
\Cref{fig:tsne} shows the effectiveness of our contrastive training approach, using t-SNE  as a dimensionality reduction tool for visualization, where we analyze the separability of different disease types in our embedding space. Specifically, it shows that text embeddings can separate well the different diseases, while image embeddings, computed as the mean of all the patches, are noisier. Differently, by selecting the 32 most similar patches to the text embedding and computing the mean of these embeddings we can still separate well the embedding space using images. This insight gives an additional motivation to use attention in our conditional model.

Furthermore, in \Cref{fig:tsne-patches}, we provide a more detailed visualization of the image embeddings after contrastive pretraining, including representative patches for each class, which highlights how the contrastive learning objective structures the latent space according to morphological similarity, grouping patches with comparable tissue organization or cellular composition while maintaining clear separation between different disease classes.

\begin{figure}[t]
    \centering
    \includegraphics[width=0.9\linewidth]{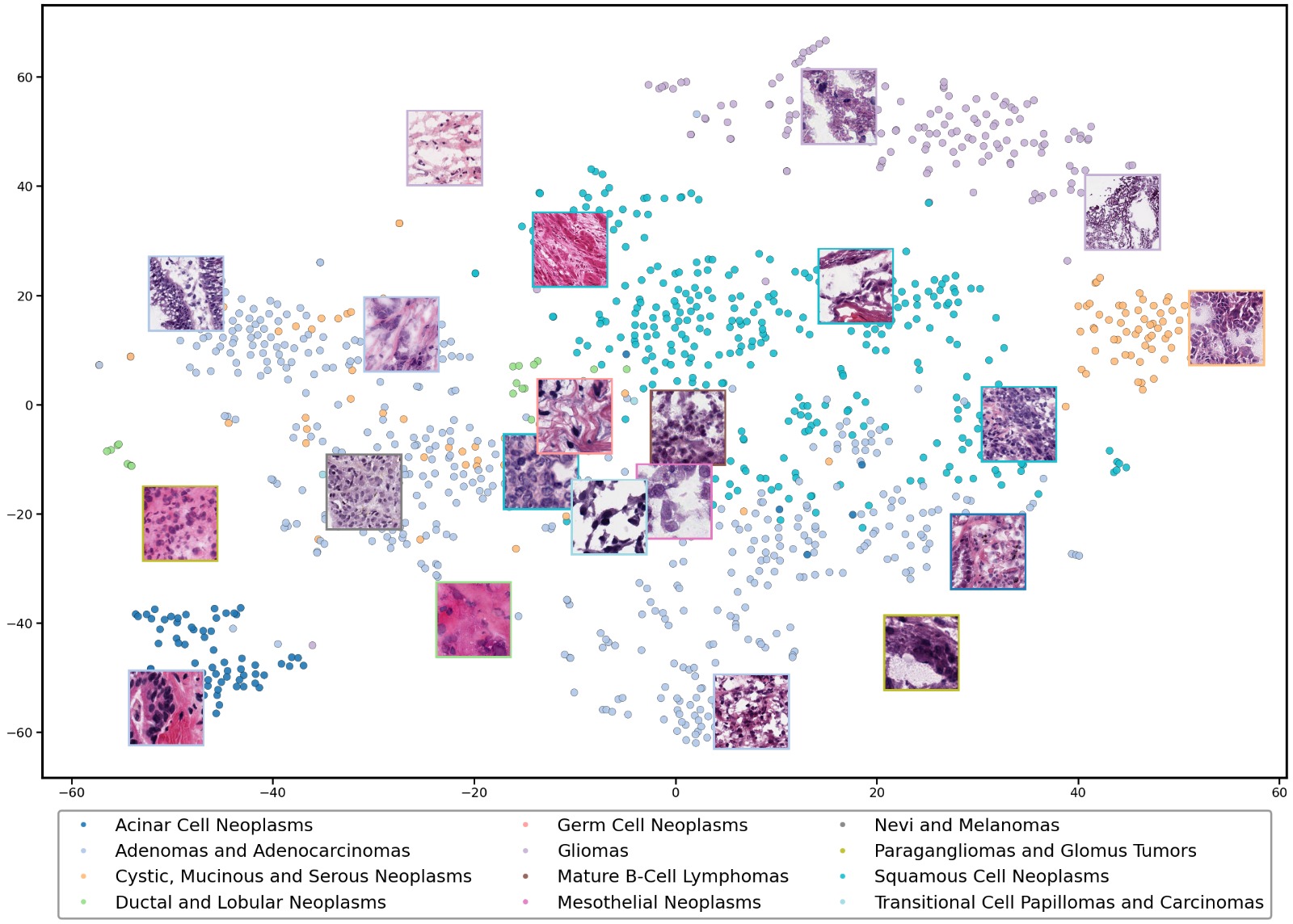}
    \caption{t-SNE visualization of image embeddings obtained after contrastive pretraining. The figure not only shows the distribution of embedding points in the latent space but also includes representative image patches for each class. This visualization highlights how the contrastive learning approach effectively separates images according to both class-specific and morphological features, producing well-separated clusters that capture meaningful relationships within the data.}
    \label{fig:tsne-patches}
\end{figure}

\subsection{Conditional WGAN-GP Results}
We train our Conditional WGAN-GP for 500 epochs using a latent dimension of 256 and the Adam optimizer with a starting learning rate of 0.0005, which decreases by a factor of 0.5 every 50 epochs.

\Cref{table:main-results} presents the results for the unsupervised and detectability metrics, comparing a vanilla WGAN-GP model without conditioning against our proposed strategy, which employs a multi-head attention mechanism to integrate multimodal information. To further assess the effectiveness of this approach, we conduct an ablation study considering three variants: single-modality conditioning (image-only or text-only) and a simple multimodal configuration that uses the mean of the text and image embeddings obtained after contrastive learning.

\begin{table}[t]
\centering
\caption{Results of unsupervised and detectability metrics (Accuracy and F1-Score), computed as the mean of 5 runs. We compare \methodName{} with a \emph{WGAN-GP} conditioned on the disease type. We denote with \emph{Image Only} and \emph{Text Only} the single-modal models trained by concatenating to the noise only the image or text embedding, while in \emph{Mean} we compute the mean of the two. Best results are in bold.}
\resizebox{\textwidth}{!}{%
\begin{tabular}{|c|>{\centering\arraybackslash}p{1.6cm}|>{\centering\arraybackslash}p{1.6cm}|>{\centering\arraybackslash}p{1.6cm}|>{\centering\arraybackslash}p{1.6cm}|>{\centering\arraybackslash}p{1.6cm}|>{\centering\arraybackslash}p{1.6cm}|>{\centering\arraybackslash}p{1.6cm}|>{\centering\arraybackslash}p{1.6cm}|}
\hline
\multirow{2}{*}{Method} & \multirow{2}{*}{\makecell{Precision \\ $\uparrow$}} & \multirow{2}{*}{\makecell{Recall \\ $\uparrow$}} & \multirow{2}{*}{\makecell{Correlation \\ $\uparrow$}} & \multirow{2}{*}{\makecell{Accuracy \\(MLP) $\downarrow$}} & \multirow{2}{*}{\makecell{F1-Score \\ (MLP) $\downarrow$}} & \multirow{2}{*}{\makecell{Accuracy \\ (LR) $\downarrow$}} & \multirow{2}{*}{\makecell{F1-Score \\ (LR) $\downarrow$}} \\
& & & & & & &\\
\hline
WGAN-GP & \textbf{0.890} & 0.778 & 0.871 & 0.927 & 0.926 & 0.851 & 0.869 \\
\hline
\methodName{}(Img Only) & 0.820 & 0.834 & \textbf{0.887} & 0.862 & 0.851 &  0.758 & 0.804 \\
\hline
\methodName{}(Text Only) & 0.885 & 0.781 & 0.886 & \textbf{0.814} & \textbf{0.804} & 0.762 & 0.804 \\
\hline
\methodName{}(Mean) & 0.822 & \textbf{0.859} & 0.880 & 0.840 & 0.829 & 0.776 & 0.816 \\
\hline
\methodName{} & 0.763 & 0.820 & 0.880 & 0.869 & 0.858 & \textbf{0.713} & \textbf{0.767} \\
\hline
\end{tabular}
}
\label{table:main-results}
\end{table}

From the unsupervised metrics, the benefits of multi-head attention are particularly evident in the detectability results, where the logistic regression classifier reaches approximately 71\% accuracy and 76\% F1 score. This indicates that attention-based conditioning produces samples that are more realistic and less easily distinguishable from real data. The model conditioned on the mean embedding, while simpler, achieves the best balance between precision and recall—maintaining a recall of around 86\% without compromising precision—suggesting that it effectively captures a broader portion of the data distribution. However, as shown in the subsequent tables, our attention-based conditioning strategy demonstrates its full potential in real-world utility evaluations. The relatively lower performance on unsupervised metrics may reflect the model’s ability to generate realistic yet novel samples that do not merely replicate the training data but instead provide meaningful variations, which is particularly valuable from both a privacy and ethical standpoint.

\begin{table}[t]
\centering
\caption{Utility evaluation with Random Forest (RF) and Multi Layer Perceptron (MLP), classifying the disease type given the gene expression. \emph{TRTR} means training only on the real data, and \emph{TSTR} involves training on the generated data. \emph{T(S+R)TR} involves using both data sources to train. The Augmentation Factor (\emph{Aug. Factor}) represents how many samples are generated for each real sample.
}
% \resizebox{\textwidth}{!}{%
\begin{tabular}{|c|c|>{\centering\arraybackslash}p{1.6cm}|>{\centering\arraybackslash}p{1.6cm}|>{\centering\arraybackslash}p{1.6cm}|>{\centering\arraybackslash}p{1.6cm}|}
\hline
\multirow{2}{*}{Training Data} & \multirow{2}{*}{Aug. Factor} & \multirow{2}{*}{\makecell{Accuracy \\ (MLP) $\uparrow$}} & \multirow{2}{*}{\makecell{F1-Score \\(MLP) $\uparrow$}} & \multirow{2}{*}{\makecell{Accuracy \\ (RF) $\uparrow$}} & \multirow{2}{*}{\makecell{F1-Score \\ (RF) $\uparrow$}} \\
& & & & & \\
\hline
TRTR & - & 0.880 & 0.870 & 0.902 & 0.882 \\
\hline
TSTR & 1 & 0.883 & 0.871 & 0.901 & 0.880 \\
\hline
TSTR & 10 & \textbf{0.897} & \textbf{0.883} & 0.908 & 0.893 \\
\hline
T(S+R)TR & 1 & 0.889 & 0.878 & 0.908 & 0.892 \\
\hline
T(S+R)TR & 10 & 0.894 & \textbf{0.883} & \textbf{0.909} & \textbf{0.894} \\
\hline
\end{tabular}
% }
\label{table:utility-evaluation}
\end{table}

We further evaluate the practical utility of our model by training Random Forest (RF) and Multi-Layer Perceptron (MLP) classifiers to predict disease type from gene expression profiles. We adopt three training strategies: \emph{TRTR} (training only on real data), \emph{TSTR} (training only on generated data), and \emph{T(S+R)TR} (training on both real and synthetic data). Results in \Cref{table:utility-evaluation} show that models trained on generated data alone achieve comparable performance to those trained on real data, and that augmenting the dataset with ten synthetic samples per real instance further improves predictive accuracy. The combined \emph{T(S+R)TR} setup consistently attains the highest scores across classifiers compared to\emph{TRTR}, suggesting that the inclusion of synthetic data can provide modest but consistent benefits for downstream prediction, even when real-world labelled data is available.

Finally, within the TSTR setting (\Cref{table:utility-ablation}), we assess the influence of the different conditioning strategies, as in the unsupervised evaluation. The attention-based model consistently achieves the highest accuracy and F1 scores across classifiers, outperforming both the unconditioned WGAN-GP and the single-modality variants. In contrast, the simple mean-based fusion performs worse than using either modality alone, confirming that effective multimodal integration requires a mechanism that can selectively weight complementary information rather than collapsing it into an averaged representation. Overall, these findings highlight the importance of attention-driven fusion in leveraging both modalities to improve the quality, realism, and downstream utility of the generated data.

\begin{table}[t]
\centering
\caption{Comparison of the utility (TSTR) of \methodName{} with the same benchmarks of \Cref{table:main-results}.}
%\resizebox{\textwidth}{!}{%
\begin{tabular}{|c|>{\centering\arraybackslash}p{1.6cm}|>{\centering\arraybackslash}p{1.6cm}|>{\centering\arraybackslash}p{1.6cm}|>{\centering\arraybackslash}p{1.6cm}|>{\centering\arraybackslash}p{1.6cm}|}
\hline
\multirow{2}{*}{Method} & \multirow{2}{*}{\makecell{Accuracy \\(MLP) $\uparrow$}} & \multirow{2}{*}{\makecell{F1-Score \\ (MLP) $\uparrow$}} & \multirow{2}{*}{\makecell{Accuracy \\ (RF) $\uparrow$}} & \multirow{2}{*}{\makecell{F1-Score \\ (RF) $\uparrow$}} \\
& & & & \\
\hline
WGAN-GP & 0.853 & 0.847 & 0.891 & 0.867 \\
\hline
\methodName{}(Img Only) & 0.863 & 0.851 & 0.894 & 0.872 \\
\hline
\methodName{}(Text Only) & 0.874 & 0.863 & \textbf{0.901} & 0.879 \\
\hline
\methodName{}(Mean) & 0.870 & 0.861 & 0.899 & \textbf{0.880} \\
\hline
\methodName{} & \textbf{0.883} & \textbf{0.871} & \textbf{0.901} & \textbf{0.880} \\
\hline
\end{tabular}
%}
\label{table:utility-ablation}
\end{table}

\subsection{Gene- and Pathway-Level Biological Coherence}
\label{sec:bio-coherence}

To further support the biological coherence of the generated transcriptomic profiles, we performed an additional gene- and pathway-level comparison between real and synthetic data on a representative disease-type contrast (\emph{Squamous Cell Neoplasms} vs \emph{Adenomas and Adenocarcinomas}) using the held-out test split (containing 83 and 102 samples for the two diseases).

\begin{table}[t]
\centering
\caption{Agreement between differentially expressed (DE) genes identified from real and generated data for the contrast \emph{Squamous Cell Neoplasms} vs \emph{Adenomas and Adenocarcinomas}.}
\begin{tabular}{|l|c|c|c|}
\hline
\textbf{Regulation} & \textbf{Top-$N$ genes} & \textbf{Overlap} & \textbf{Jaccard similarity} \\
\hline
Up-regulated   & 100 & 89 & 0.80 \\
\hline
Down-regulated & 100 & 86 & 0.75 \\
\hline
\end{tabular}
\label{table:de-overlap}
\end{table}

\begin{figure}[t]
    \centering
    \includegraphics[width=\linewidth]{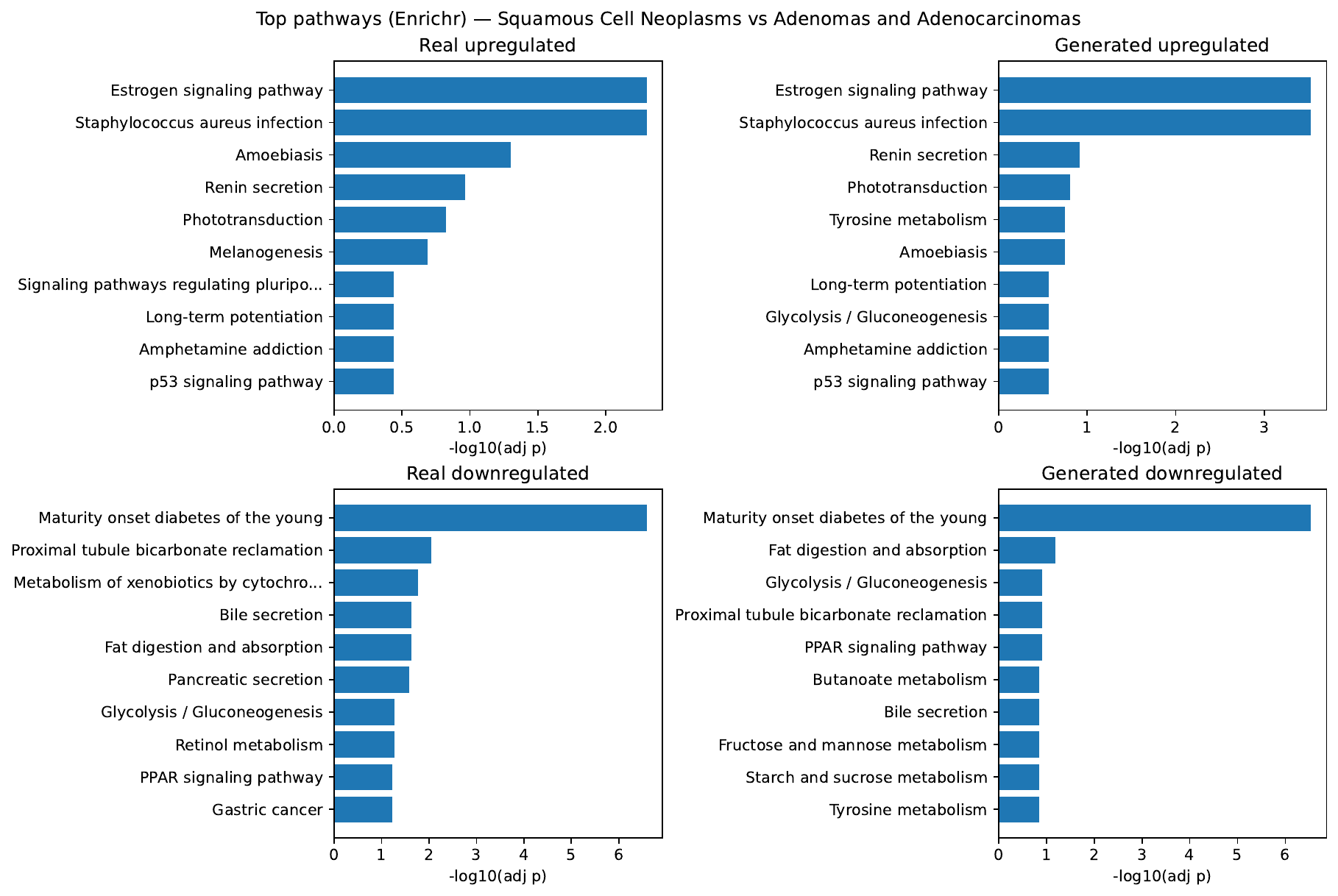}
    \caption{Top enriched pathways for the contrast \emph{Squamous Cell Neoplasms} vs \emph{Adenomas and Adenocarcinomas}, computed from the top-$100$ up-regulated (top row) and down-regulated (bottom row) genes. Results are shown for real data (left) and synthetic profiles generated by \methodName{} (right). Bar lengths represent pathway significance as $-\log_{10}$ adjusted p-values, highlighting a strong agreement between real and generated data.}
    \label{fig:top-pathways}
\end{figure}

\paragraph{Differential expression consistency.}
For each gene, we compared the two disease groups using a two-sided non-parametric Mann--Whitney U test on log-transformed expression values ($\log_2(\mathrm{FPKM}+1)$), with multiple-testing correction performed using the Benjamini--Hochberg procedure. We selected the top-$100$ most deregulated genes separately for up- and down-regulation and repeated the same analysis on synthetic profiles generated by \methodName{} for the same disease contrast. A quantitative summary of the agreement between real and generated data is reported in \Cref{table:de-overlap}, showing a strong overlap for both up-regulated (89/100, Jaccard $=0.80$) and down-regulated genes (86/100, Jaccard $=0.75$). These results indicate that disease-associated transcriptional differences are largely preserved in the generated profiles.

\paragraph{Pathway-level agreement.}
To assess whether this gene-level consistency translates into coherent biological processes, we performed enrichment analysis on the derived gene sets using a standard pathway gene-set library. The comparison of the top enriched pathways obtained from real and generated data reveals a substantial overlap among the top-$10$ pathways (8/10, Jaccard $=0.67$). Representative results for both up- and down-regulated genes are reported in \Cref{fig:top-pathways}, highlighting that synthetic data capture pathway-level trends consistent with those observed in real samples.

\subsection{Explainability Analysis}

\begin{figure}[t]
    \centering
    \includegraphics[width=\linewidth]{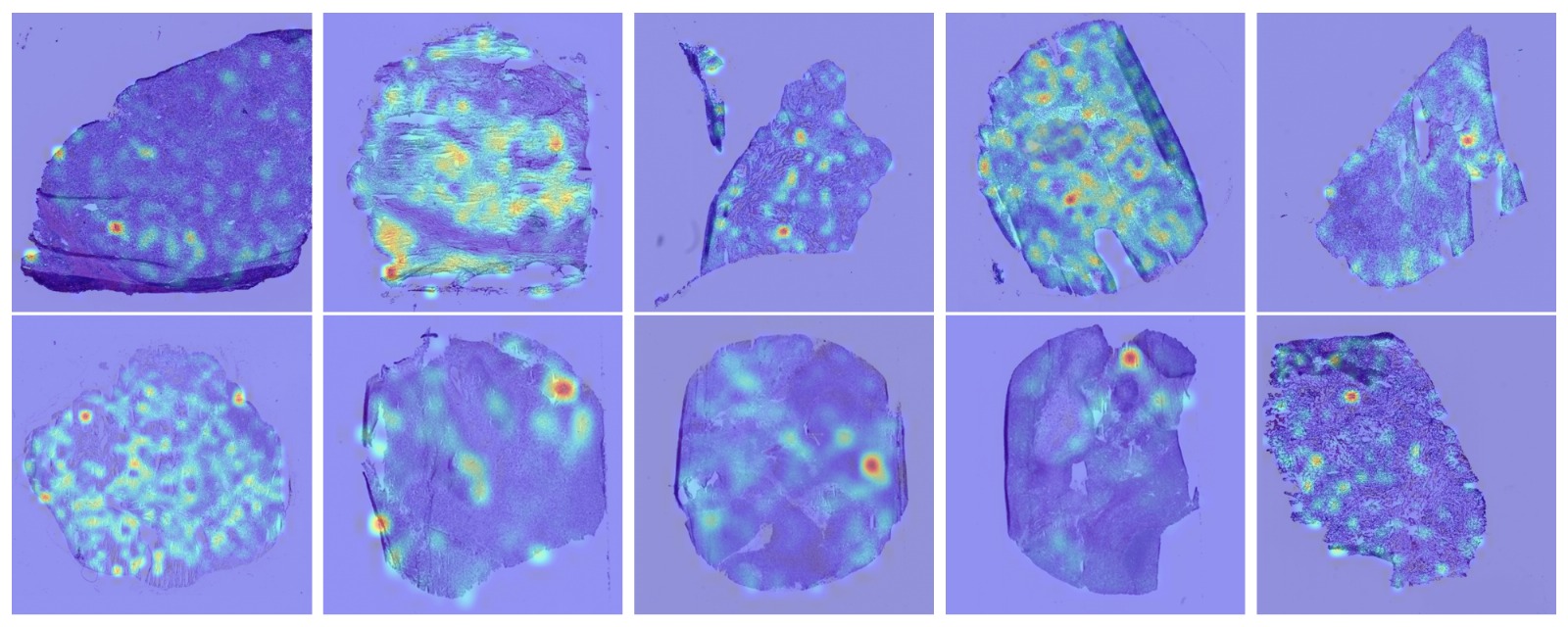}
    \caption{Attention maps generated by the multi-head attention mechanism for ten representative histopathology slides. Warmer colors (yellow–red) highlight tissue regions assigned higher attention weights, indicating greater influence on the synthesis of the corresponding gene expression profiles. These visualizations illustrate the interpretability of \methodName{}, revealing how the model selectively focuses on biologically relevant structures within each slide when generating molecular data.
    % \textcolor{red}{Attention maps for a selection of ten randomly sampled image patches. Visualization of attention weights from the multi-head attention mechanism across ten randomly sampled histopathology patches. Warmer colors (yellow-red hotspots) indicate the regions within each tissue patch that contributed most strongly to the generator’s synthesis of the corresponding gene expression profile.}
    }
    \label{fig:attention-maps}
\end{figure}

To demonstrate the intrinsic interpretability of our multimodal architecture, we explore the generator’s inner workings through its multi-head attention mechanism. As visualized in \Cref{fig:attention-maps}, the attention weights provide a direct visualization of how the model integrates histopathological context when synthesizing gene expression profiles. This analysis provides insight into the model’s decision-making, highlighting the regions of the WSI that have contributed most to the generation process. We examined the attention weights for ten randomly selected tissue slides to assess how the model prioritizes different patches. The resulting maps highlight distinct hotspots (shown in warm colors), indicating the regions that the generator relies on most heavily. Importantly, this confirms that the attention mechanism can selectively identify biologically relevant tissue structures that drive gene expression. By establishing a direct, traceable link between microscopic image features and molecular outputs, this approach enhances the interpretability and transparency of our generative framework.

\subsection{Robustness to Common Histopathology Artifacts}

\begin{figure}[t]
    \centering
    \includegraphics[width=\linewidth]{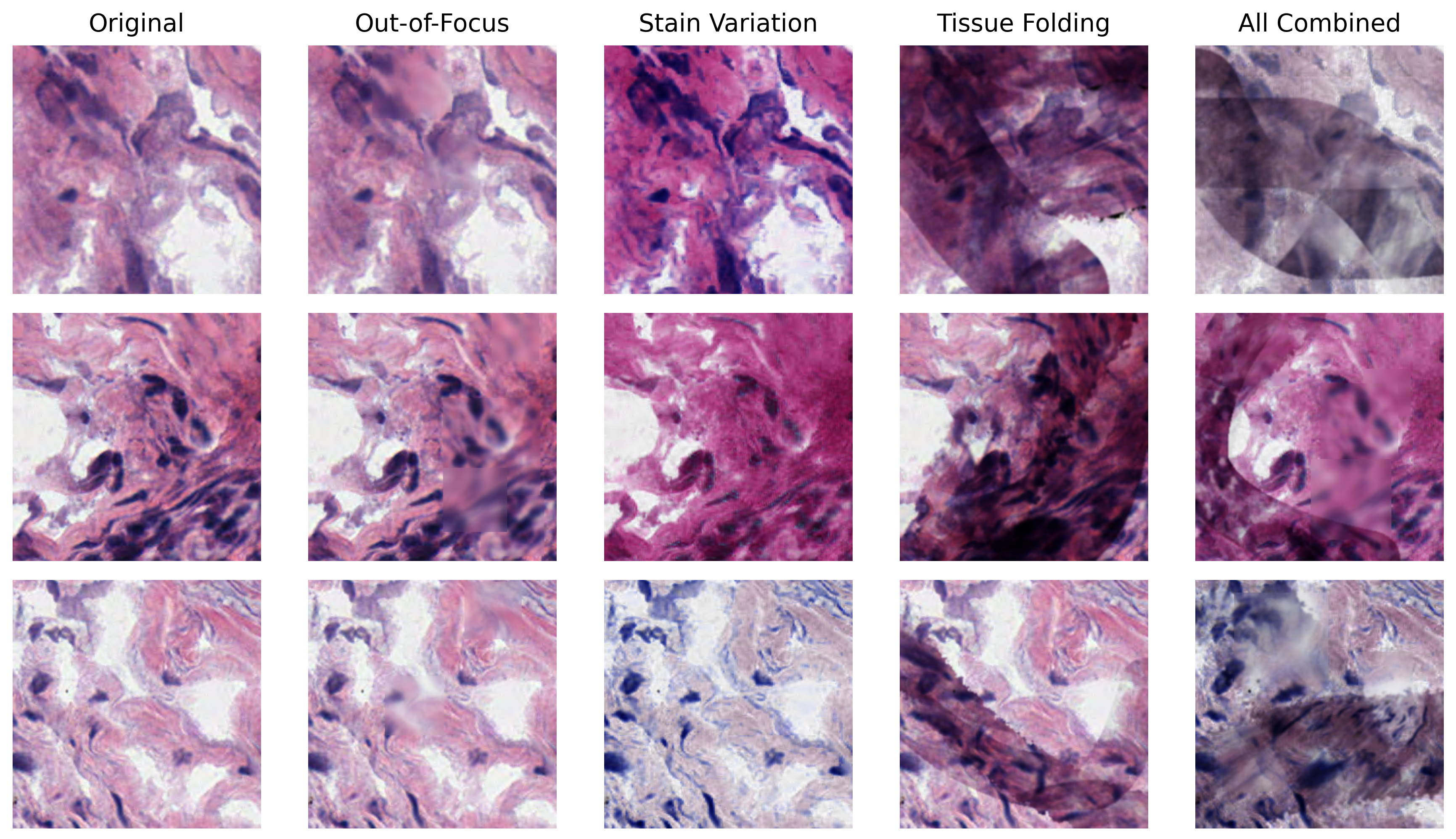}
    \caption{Examples of simulated histopathology image artifacts used in the robustness analysis. From left to right: no artifacts (clean image), out-of-focus blur, stain variation, tissue folding, and combined artifacts (out-of-focus, stain variation, and tissue folding). Artifacts are synthetically applied at the patch level to mimic common real-world degradations observed in clinical whole-slide images.}
    \label{fig:tsne-patches}
\end{figure}

\begin{table}[t]
\centering
\caption{Robustness evaluation of \methodName{} under common histopathology image artifacts: out-of-focus (OOF), stain variation (SV) and tissue folding (TF). Results are the mean of 5 independent runs.}
\resizebox{\textwidth}{!}{%
\begin{tabular}{|>{\centering\arraybackslash}p{0.8cm}|>{\centering\arraybackslash}p{0.8cm}|>{\centering\arraybackslash}p{0.8cm}|>{\centering\arraybackslash}p{1.6cm}|>{\centering\arraybackslash}p{1.6cm}|>{\centering\arraybackslash}p{1.6cm}|>{\centering\arraybackslash}p{1.6cm}|>{\centering\arraybackslash}p{1.6cm}|>{\centering\arraybackslash}p{1.6cm}|>{\centering\arraybackslash}p{1.6cm}|>{\centering\arraybackslash}p{1.6cm}|}

\hline
\multicolumn{3}{|c|}{\textbf{Artifact}} &
\multicolumn{3}{c|}{\textbf{Unsupervised}} &
\multicolumn{2}{c|}{\textbf{Detectability}} &
\multicolumn{2}{c|}{\textbf{Utility}} \\

\hline
\multirow{2}{*}{OOF} & 
\multirow{2}{*}{SV} &  
\multirow{2}{*}{TF} & 
\multirow{2}{*}{\makecell{Precision \\ $\uparrow$}} & 
\multirow{2}{*}{\makecell{Recall \\ $\uparrow$}} & 
\multirow{2}{*}{\makecell{Correlation \\ 
$\uparrow$}} & \multirow{2}{*}{\makecell{Accuracy \\(LR) $\downarrow$}} & 
\multirow{2}{*}{\makecell{F1-Score \\ (LR) $\downarrow$}} & 
\multirow{2}{*}{\makecell{Accuracy \\ (RF) $\downarrow$}} & 
\multirow{2}{*}{\makecell{F1-Score \\ (RF) $\downarrow$}}\\
& & & & & & & & & \\
\hline
& & & 0.763 & 0.820 & 0.880 & 0.713 & 0.767 & 0.901 & 0.880 \\
\hline
\checkmark & & & 0.806 & 0.778 & 0.875 & 0.757 & 0.803 & 0.900 & 0.878 \\
\hline
& \checkmark & & 0.812 & 0.755 & 0.864 & 0.742 & 0.788 & 0.899 & 0.876 \\
 \hline
& & \checkmark & 0.812 & 0.749 & 0.873 & 0.729 & 0.773 & 0.900 & 0.878 \\
\hline
\checkmark & \checkmark & \checkmark & 0.841 & 0.702 & 0.851 & 0.707 & 0.751 & 0.899 & 0.876 \\
\hline
\end{tabular}
}
\label{table:robustness-wsi}
\end{table}

In real-world clinical settings, histopathology whole-slide images are often affected by acquisition and preparation artifacts, such as out-of-focus regions, staining variability, or tissue folding. These artifacts may introduce noise or distort morphological patterns, potentially impacting the reliability of downstream models. To assess the robustness of \methodName{} under such non-ideal conditions, we conducted a robustness analysis by evaluating the model in the presence of common histopathological artifacts.

Specifically, we considered three widely observed artifact types: out-of-focus blur, staining variations, and tissue folding. Each artifact was synthetically introduced at the image level prior to feature extraction, while keeping the clinical metadata and gene expression profiles unchanged. For stain variation and tissue folding, we followed the same procedures of \cite{WANG202154} and applied them to each patch of the WSI independently. To simulate, instead, out-of-focus regions, we apply a Gaussian blur to random regions of the patches. \Cref{fig:artifacts} reports some visualizations of the simulated artifacts.

We designed five experimental settings: \emph{(i)} no artifacts (i.e. clean images), \emph{(ii)} out-of-focus only, \emph{(iii)} stain variation only, \emph{(iv)} tissue folding only, and \emph{(v)} a combined scenario where all three artifacts are present simultaneously. For each setting, we evaluated \methodName{} following exactly the same protocol adopted in \Cref{table:main-results,table:utility-evaluation}, evaluating the model with unsupervised metrics, detectability and utility. All metrics were computed as the mean over five independent runs, ensuring direct comparability with the main experimental results.

\Cref{table:robustness-wsi} reports the results and indicates that \methodName{} is highly robust to moderate image degradations. Across all artifact configurations, performance remains largely stable, with only marginal variations compared to the artifact-free baseline. A closer inspection of \Cref{table:robustness-wsi} highlights a few consistent trends. Precision and recall exhibit a trade-off across different artifact configurations, with improvements in one metric typically accompanied by slight decreases in the other, indicating minor shifts in distribution coverage rather than a loss of overall fidelity. Among individual artifacts, tissue folding has the strongest impact, although the degradation remains limited. As expected, the configuration combining all three artifacts yields the lowest performance. Nevertheless, even in this most challenging setting, results remain superior to those obtained with the unconditioned WGAN-GP reported in \Cref{table:main-results,table:utility-evaluation}. This suggests that histopathology images remain informative even when moderately corrupted, and that the multimodal design of \methodName{}, leveraging both visual and clinical textual information, provides additional robustness under realistic imaging conditions. These findings support the applicability of \methodName{} in realistic clinical scenarios, where image imperfections are unavoidable, and further strengthen its potential for deployment in practical biomedical research pipelines.

\section{Conclusions}

This work introduced a multimodal generative framework that conditions gene expression synthesis on both clinical metadata and histopathology images. Our aim was to generate gene expression profiles that are not only statistically realistic but also clinically coherent for real-world applications. Our results demonstrate that the model performs strongly across standard generative evaluation metrics, producing synthetic gene expression profiles that closely match real data distributions. Compared to existing generative approaches for gene expression profiles, our model not only improves realism and utility in downstream predictive tasks but also introduces an interpretable conditioning mechanism providing clear insights into which regions of the images are most influential in producing specific synthesized profiles. This interpretability is particularly valuable, as it opens the door for future studies to generate pathological gene expression profiles and investigate the morphological features most relevant to their manifestation. 

From a clinical and translational standpoint, \methodName{} is not intended to replace molecular profiling, but rather to support research and development workflows in which gene expression data are scarce, costly, or constrained by privacy considerations. By enabling the generation of biologically coherent transcriptomic profiles conditioned on routinely available data, namely histopathology images and clinical metadata, the proposed framework facilitates in silico experimentation, data augmentation for downstream models, and exploratory analyses without requiring additional wet-lab assays. Moreover, the intrinsic interpretability of the multimodal attention mechanism provides a principled way to relate tissue morphology to molecular variation, supporting hypothesis generation in computational pathology and precision medicine settings.

Future work will focus on benchmarking \methodName{} against additional generative models and further leveraging its multimodal foundation. A promising direction is \emph{bidirectional synthesis}: generating virtual histopathology images from gene expression inputs and studying how changes in one modality affect the other. This approach enables in silico experiments that are privacy-preserving and cost-effective, offering new opportunities to explore the molecular consequences of patient context and tissue morphology.

\section*{Acknowledgements}
Model training and testing were possible thanks to the HPC grant from by the Ministry of Education, Youth and Sports of the Czech Republic through the e-INFRA CZ (ID:90254)

%
% the environments 'definition', 'lemma', 'proposition', 'corollary',
% 'remark', and 'example' are defined in the LLNCS documentclass as well.
%

%
% ---- Bibliography ----
%
% BibTeX users should specify bibliography style 'splncs04'.
% References will then be sorted and formatted in the correct style.
%
\bibliographystyle{splncs04}
\bibliography{bibliography_CIBB_file}
%
% \begin{thebibliography}{8}
% \bibitem{ref_article1}
% Author, F.: Article title. Journal \textbf{2}(5), 99--110 (2016)

% \bibitem{ref_lncs1}
% Author, F., Author, S.: Title of a proceedings paper. In: Editor,
% F., Editor, S. (eds.) CONFERENCE 2016, LNCS, vol. 9999, pp. 1--13.
% Springer, Heidelberg (2016). \doi{10.10007/1234567890}

% \bibitem{ref_book1}
% Author, F., Author, S., Author, T.: Book title. 2nd edn. Publisher,
% Location (1999)

% \bibitem{ref_proc1}
% Author, A.-B.: Contribution title. In: 9th International Proceedings
% on Proceedings, pp. 1--2. Publisher, Location (2010)

% \bibitem{ref_url1}
% LNCS Homepage, \url{http://www.springer.com/lncs}, last accessed 2023/10/25
% \end{thebibliography}
\end{document}